\documentclass[twoside,11pt]{article}

\usepackage{jmlr2e}
\usepackage{enumitem}
\usepackage{xcolor}
\usepackage{blindtext}

\ShortHeadings{AutoOED: Automated Optimal Experiment Design Platform}{Tian et al.}
\firstpageno{1}

\begin{document}

\title{AutoOED: Automated Optimal Experiment Design Platform}

\author{\name Yunsheng Tian \email yunsheng@csail.mit.edu \\
       \AND
       \name Mina Konaković Luković \email mina@csail.mit.edu \\
       \AND
       \name Timothy Erps \email terps@csail.mit.edu \\
       \AND
       \name Michael Foshey \email mfoshey@csail.mit.edu \\
       \AND
       \name Wojciech Matusik \email wojciech@csail.mit.edu \\
       \addr Computer Science \& Artificial Intelligence Lab\\
       Massachusetts Institute of Technology\\
       Cambridge, MA 02139, USA}


\maketitle

\begin{abstract}
We present AutoOED, an Optimal Experiment Design platform powered with automated machine learning to accelerate the discovery of optimal solutions. The platform solves multi-objective optimization problems in time- and data-efficient manner by automatically guiding the design of experiments to be evaluated. To automate the optimization process, we implement several multi-objective Bayesian optimization algorithms with state-of-the-art performance. AutoOED is open-source and written in Python. The codebase is modular, facilitating extensions and tailoring the code, serving as a testbed for machine learning researchers to easily develop and evaluate their own multi-objective Bayesian optimization algorithms. An intuitive graphical user interface (GUI) is provided to visualize and guide the experiments for users with little or no experience with coding, machine learning, or optimization. Furthermore, a distributed system is integrated to enable parallelized experimental evaluations by independent workers in remote locations. The platform is available at \url{https://autooed.org}.
\end{abstract}

\begin{keywords}
  multi-objective optimization, Bayesian optimization, Pareto optimality, optimal experiment design, design of experiments
\end{keywords}

\section{Introduction}
\label{sec:introduction}
Optimal Experiment Design (OED) problems in science and engineering often require satisfying several conflicting objectives simultaneously. These problems aim to solve a multi-objective optimization system and discover a set of optimal solutions, called Pareto optimal. Furthermore, the objectives are typically black-box functions whose evaluations are time-consuming and costly (e.g., evaluations can either measure real experiments or run expensive numerical simulations). Thus, the budget that determines the number of experiments can be heavily constrained. Hence, an efficient strategy for guiding the experimental design towards Pareto-optimal solutions is necessary. Recent advances in machine learning algorithms have facilitated optimization of various design problems, including chemical design~\citep{griffiths2017constrained}, material design~\citep{zhang2020bayesian}, battery design~\citep{attia2020closed}, clinical drug trials~\citep{yu2019drugs}, resource allocation~\citep{Wu2013}, environmental monitoring~\citep{Marchant2012environment}, 
robotics~\citep{Martinez-Cantin2009}. A machine learning concept that enables automatic guidance of the design process is Bayesian optimization~\citep{BOsurvey}. This concept is extensively studied in the machine learning and optimization community from a theoretical aspect and in the single-objective case. However, its potential applications in other fields with multi-objective problems are still not widely explored due to the lack of intuitive, easy to use, and open-source software. BoTorch~\citep{botorch} is a recent library for Bayesian optimization, nevertheless, it lacks a GUI, does not  support distributed usage, and is targeted for experts in coding. Auto-QChem~\citep{shields2021bayesian} is a recent open-source software for experiment design in chemistry, but it does not optimize for multiple objectives and does not apply to fields other than chemistry. 


AutoOED is an open-source platform for multi-objective optimization problems with a restricted budget of test samples. The platform automatically guides the design of experiments to be evaluated and quickly leads to the best performing designs. AutoOED is based on the concept of multi-objective Bayesian optimization (MOBO) and is entirely implemented in Python. The key features of this platform include:
\begin{itemize}[leftmargin=*,noitemsep,topsep=2pt]
\item \textbf{Intuitive GUI}: An easy-to-use graphical user interface (GUI) is provided to directly visualize and guide the optimization progress and facilitate the operation for users with little or no experience with coding, optimization, or machine learning. The GUI is built using the Tkinter package with a model-view-controller (MVC) architecture.
\item \textbf{Modular structure}:  A highly modularized codebase and built-in visualization enable easy extensions and replacements of MOBO algorithm components. The platform can serve as a testbed for machine learning researchers to easily develop and evaluate their own MOBO algorithms.
\item \textbf{Data-efficient experimentation}: As the platform aims to solve 
problems with expensive to evaluate or black-box objective functions, the number of experiments is limited to an order of several dozens. The platform employs an optimization strategy that rapidly advances the Pareto front with a small set of evaluated experiments. 
\item \textbf{Sequential and batch evaluations}: A standard feature supports evaluating a single experiment in each optimization iteration. To further reduce the optimization time, the platform parallelizes the evaluation process by enabling synchronous and asynchronous batch evaluations. Asynchronous batch evaluations are instrumental when multiple workers are running experiments, but their evaluations drastically vary in time.
\item \textbf{Automation of experiment design}: The platform is designed for straightforward integration into an automatic experiment design optimization pipeline. The pipeline may include both physical experiments (see example at \url{https://tinyurl.com/autooed-phys}) and an expensive simulation setup (see example at \url{https://tinyurl.com/autooed-sim}). 

\item \textbf{Distributed usage}: Besides the standard single-user version, to facilitate the platform’s use for a single design process by several users, we provide a distributed team version based on a centralized database. The database shares the optimization status to different roles in the experiment team (managers, scientists, technicians) and provides them with different tools for controlling and contributing to the experimental design process.
\end{itemize}

\section{Platform Overview}
\label{sec:overview}
To automate the optimization process and guide the experiments, AutoOED is based on the principles and algorithms of multi-objective Bayesian optimization (MOBO). 
MOBO is a data-driven approach that attempts to learn the black-box objective functions from available data and find Pareto-optimal solutions in a data-efficient manner.
AutoOED outputs the whole palette of designs on the Pareto front, enabling the user to choose the trade-off most suited for their application. In addition to the set of optimal solutions, our platform's final product is the learned prediction models of the unknown objectives. The prediction provides the users with more insights into the potential outcomes of experiments. It helps them better understand the optimization problem to make informed decisions and guide the optimization process towards their preference.

\begin{figure}[ht!]
    \centering
    \vspace{-5pt}
    \includegraphics[width=0.8\textwidth]{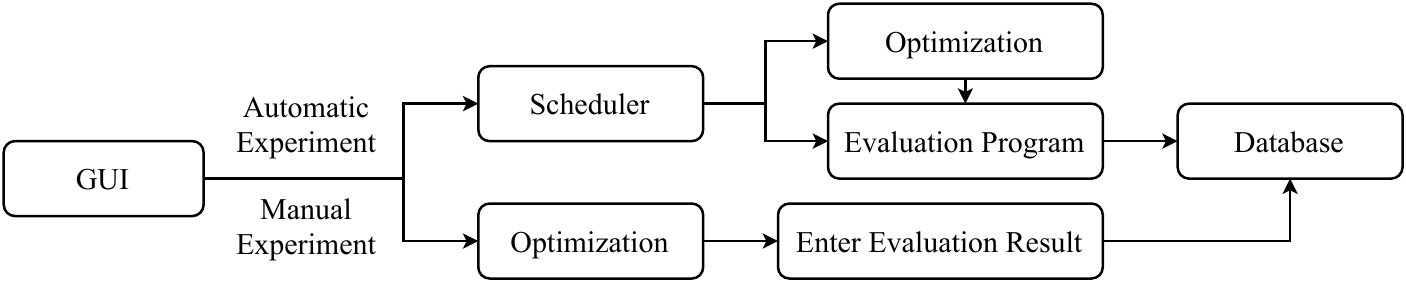}
    \vspace{-5pt}
    \caption{The overall workflow of the platform design.}
    \vspace{-5pt}
    \label{fig:workflow}
\end{figure}

As a full-stack software, the overall workflow of AutoOED from the front-end to the back-end is presented in Figure~\ref{fig:workflow}. 
Given an optimization problem, if the evaluation program is available, AutoOED can automatically guide the experimentation by alternating between optimization and evaluation. In this case, an optimization process starts first. The scheduler then orders evaluations in an asynchronous way, and the optimization restarts once the evaluations from the currently requested batch are all done. This process repeats until some stopping criterion is met and the evaluated results are automatically recorded in the database. Otherwise, if the evaluation program does not exist, users can still manually do the evaluations and enter results into the database.

\subsection{Modular Code Structure}
\begin{figure}[ht!]
    \centering
    \vspace{-5pt}
    \includegraphics[width=1.0\textwidth]{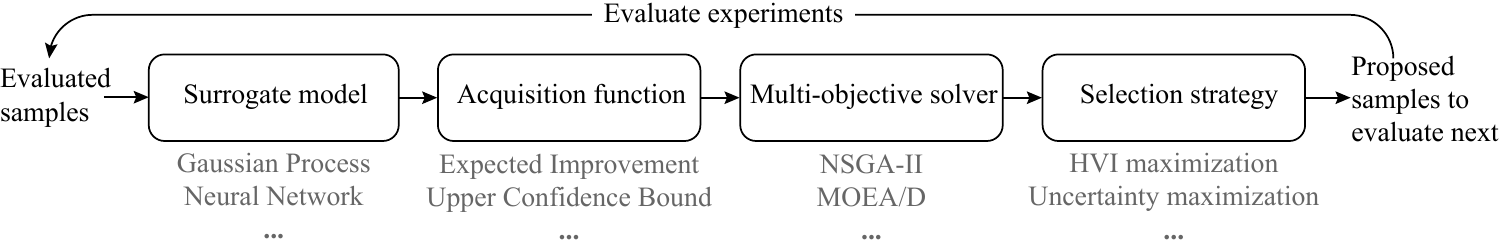}
    \vspace{-20pt}
    \caption{Algorithmic pipeline and core modules. The platform follows the multi-objective Bayesian optimization pipeline. A user can choose an approach for each module.}
    \vspace{-20pt}
    \label{fig:modules}
\end{figure}

Multi-objective Bayesian optimization typically consists of four core components: (\emph{i}) an inexpensive surrogate model for the black-box objective functions; (\emph{ii}) an acquisition function that defines sampling from the surrogate model and trade-off between exploration and exploitation of the design space; (\emph{iii}) a multi-objective optimization solver to approximate the Pareto set and front; (\emph{iv}) a selection strategy that proposes a single or a batch of experiments to evaluate next.  These four components (see Figure~\ref{fig:modules}) are implemented as core and independent building blocks of the AutoOEDs, making it highly modularized and easy to develop new algorithms and modules. A list of supported surrogate models, acquisition functions, solvers, and selection strategies can be found at the API reference section of our documentation.

\subsection{Supported Algorithms and Features}
We provide several popular and state-of-the-art multi-objective Bayesian optimization methods, including DGEMO~\citep{dgemo}, TSEMO~\citep{bradford2018efficient}, USeMO~\citep{belakaria2020uncertainty}, ParEGO~\citep{knowles2006parego}. To the best of our knowledge, DGEMO exhibits state-of-the-art performance for data-efficient, multi-objective problems with batch evaluations. Implemented algorithms also include NSGA-II~\citep{deb2002fast} and MOEA/D~\citep{4358754}, as they are some of the most widely used multi-objective evolutionary algorithms. 
Users  can  select  an  algorithm from this library that best fits the characteristics of their physical system or the optimization goals.

Additional features of the platform support: continuous, discrete, binary, categorical, and mixed design variables; linear and non-linear constraints handling in the design space; 
sequential and batch evaluations; synchronous and asynchronous batch evaluations. 
In the synchronous case, the algorithm starts a new iteration when all the requested evaluations from one batch are completed. The asynchronous batch evaluations are instrumental when evaluations drastically vary in time. To be more time-efficient, the asynchronous approach updates a model and requests a new evaluation as soon as one batch sample is completed. Our platform is especially beneficial for applications in areas such as finance, organization optimization, materials science, life sciences, chemistry, robotics, and the pharmaceutical industry, where the experiments are long (e.g., days or weeks) but can easily be carried out in parallel. 


\subsection{Graphical User Interface (GUI) and Distributed Usage} 
The GUI guides the user through a set of simple steps to configure the problem, such as the number of design and performance parameters, the parameter ranges and constraints, parallelization settings, and selection of the optimization algorithm. 
Other functional features include real-time display of design and performance space, exporting of the database and statistics, and support for linking custom evaluation scripts.
We provide a manual GUI for entering the experiment data and example scripts that showcase how to interface the worker with simulation scripts (e.g., MATLAB structural analysis toolbox, see \url{https://tinyurl.com/autooed-sim}) or automated robotic platforms that perform experiments (see examples at \url{https://tinyurl.com/autooed-phys}). 



For additional flexibility, a single experiment can be managed by several users from different computers. A distributed team version is based on a centralized database, providing secure data storage and retrieval. Users can be assigned different roles (scientist, technician, manager), 
and they will have different access and tools for controlling the experimental design process. This version helps the team collaborate more efficiently across the globe.

\section{Concluding Remarks}
\label{sec:conclusion}
AutoOED is released under the MIT license as a free and open-source platform that can be obtained at \url{https://github.com/yunshengtian/AutoOED}. This paper presents only a brief overview of the platform and its features. Comprehensive documentation and examples are available at \url{https://autooed.readthedocs.io}. The official webpage, as well as the past and ongoing projects using AutoOED, can be found at \url{https://autooed.org}. 

\acks{We would like to acknowledge support and feedback for this project
from Konica Minolta. M. K. Lukovi\'c would like to acknowledge support from the Schmidt Science Fellowship.}

\bibliography{references}

\end{document}